\documentclass[letterpaper]{article}
\PassOptionsToPackage{table}{xcolor}
\usepackage[preprint]{aaai2027}
\usepackage[hyphens]{url}
\usepackage{graphicx}
\usepackage{natbib}
\usepackage{caption}
\usepackage{placeins}
\usepackage{amsmath}
\usepackage{amssymb}
\usepackage{booktabs}
\usepackage{multirow}
\usepackage{array}
\usepackage{microtype}
\usepackage{needspace}
\usepackage[most]{tcolorbox}

\newtcolorbox{promptbox}{
    colback=teal!5,
    colframe=teal!60!black,
    boxrule=0.55pt,
    arc=0pt,
    left=4pt,
    right=4pt,
    top=3pt,
    bottom=3pt,
    boxsep=0pt,
    before skip=0.45em,
    after skip=0.65em,
    fontupper=\small
}

\newcommand{\method}{SPAR}
\newcommand{\ruler}{RULER}
\newcommand{\nolima}{NoLiMa}
\newcommand{\kl}{\mathrm{KL}}
\newcommand{\clm}{\mathcal{L}_{\mathrm{CLM}}}
\newcommand{\linv}{\mathcal{L}_{\mathrm{inv}}}

\title{Learning When Not to Listen: Selective Anti-Interference Pretraining for Language Models}

\author{
    Jinchang Zhu\textsuperscript{1,a}\equalcontrib,
    Haowei He\textsuperscript{2}\equalcontrib,
    Yi Ding\textsuperscript{1},
    Rong Fu\textsuperscript{3},
    Nie Xiaojian\textsuperscript{1},\\
    Shuangyong Song\textsuperscript{2},
    Zhongjiang He\textsuperscript{1}\corresponding,
    Menglin Yang\textsuperscript{1,b}\corresponding
}
\affiliations{
    \textsuperscript{1}The Hong Kong University of Science and Technology (Guangzhou)\\
    \textsuperscript{2}Institute of Artificial Intelligence (TeleAI), China Telecom\\
    \textsuperscript{3}University of Macau\\
    \textsuperscript{a}\texttt{jzhu997@connect.hkust-gz.edu.cn}
    \quad
    \textsuperscript{b}\texttt{menglinyang@hkust-gz.edu.cn}
}

\begin{document}

\maketitle

\begin{abstract}
Language models can over-condition on irrelevant preceding text: predictions already supported by local context may still change when distant, unrelated prefix tokens are perturbed. This interference is especially consequential in long, packed, or distractor-heavy contexts, where useful evidence and irrelevant spans coexist. We propose Selective Prefix Anti-Interference Regularization (\method), a pretraining objective for selective anti-interference. \method\ runs the original sequence and a corrupt-prefix input in which only the far prefix is changed, then uses a short-context sufficiency gate and a gated KL objective to stabilize locally supported suffix predictions. The gate operationalizes a model-based estimate of whether the far prefix supplies additional information about the target token. Mechanism analyses show that the gate identifies locally sufficient tokens and sharply reduces prefix sensitivity on gate-selected suffix tokens. In continued training on pretrained base models, \method\ improves \ruler\ across Qwen2.5-0.5B, Qwen2.5-3B, Llama-3.2-1B, Llama-3.1-8B, and GPT2-XL under equal counted training compute; pretraining experiments further show gains on both \ruler\ and \nolima. These results show that selective anti-interference is an effective objective-level signal for robust context use.

\end{abstract}

\section{Introduction}

Language models are trained to condition on previous tokens. Robust context use requires selecting the relevant evidence while resisting irrelevant preceding text. This anti-interference problem becomes especially visible in long, packed, or distractor-heavy contexts, where useful evidence and distractors coexist in the same window. Long-context evaluations repeatedly expose this selection challenge: models use long inputs unevenly \citep{liu2024lost}, lose accuracy as synthetic long-context tasks become more structured \citep{hsieh2024ruler}, and struggle when retrieval requires latent association beyond literal overlap \citep{modarressi2025nolima}.

The training objective behind this behavior is under-specified. Standard causal language modeling maximizes next-token likelihood under the observed full prefix, so all preceding tokens enter through a single likelihood target. This signal teaches the model to exploit context when it helps prediction, while the separation between useful evidence and irrelevant far-prefix variation is left implicit. If a token is already predictable from its local suffix, perturbing distant unrelated text should leave the model's prediction nearly unchanged. Ordinary CLM lacks an explicit invariance term for this stability.

We introduce Selective Prefix Anti-Interference Regularization (\method), a pretraining objective that teaches this selective stability. Given a training sequence, \method\ preserves the local suffix and corrupts only the far prefix in a second input. The model performs an original-sequence forward pass and a corrupt-prefix forward pass. A short-context sufficiency gate selects suffix tokens that are already locally predictable, and a gated KL loss stabilizes the predictions for those selected tokens under far-prefix corruption. The objective gives the model an explicit anti-interference signal alongside ordinary CLM.

The central idea is simple: CLM teaches models to use context; \method{} teaches them which context should matter less for locally sufficient targets. The ordinary CLM objective remains responsible for learning predictive evidence from the observed prefix. The gated invariance term adds a second behavior at positions where the recent suffix already supports the target and the far prefix contributes little additional likelihood. This design lets long-range evidence remain useful at high-gap positions while giving locally determined predictions a direct stability target.

\begin{figure*}[t]
    \centering
    \includegraphics[width=\textwidth]{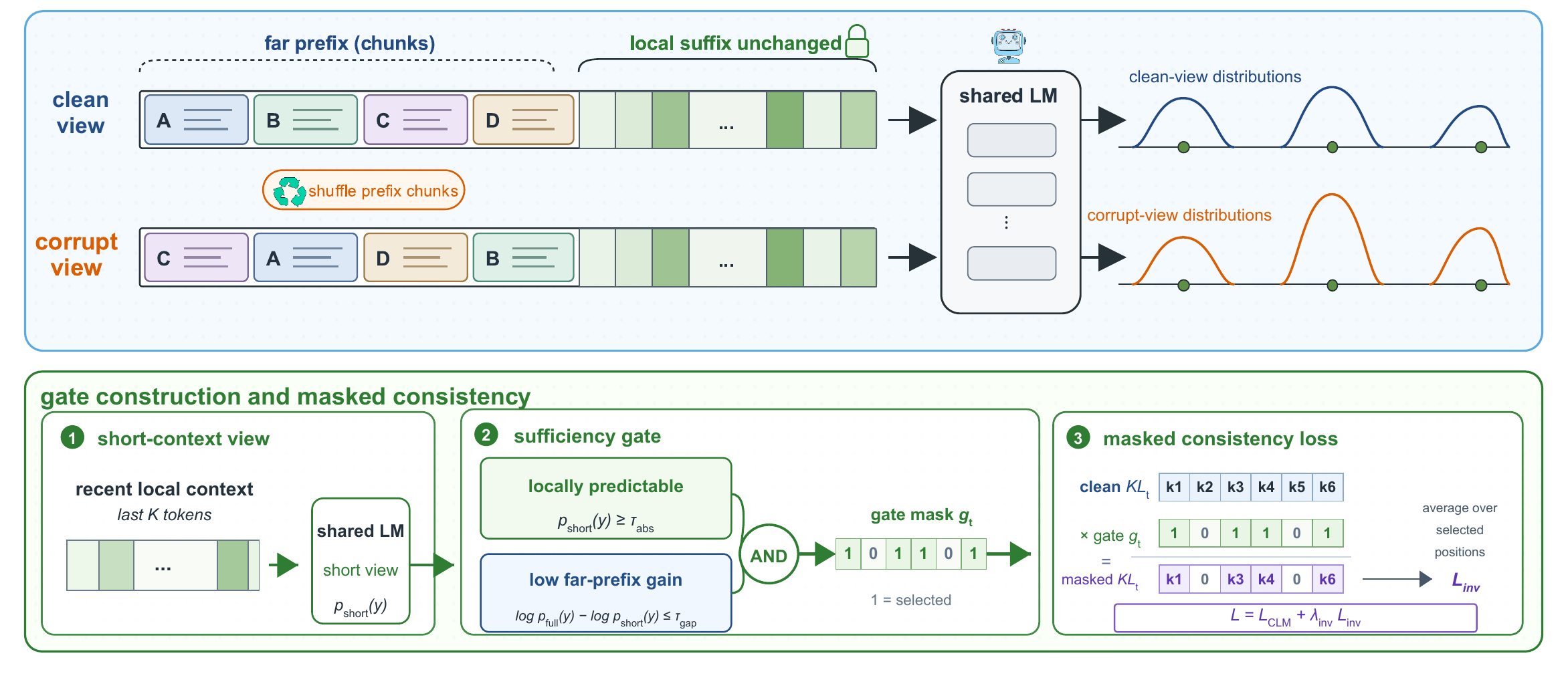}
    \caption{\method\ forms a corrupt-prefix view by shuffling chunks only in the far prefix while preserving the local suffix token for token. The short-context gate selects suffix tokens with high short-context likelihood and a small full--short likelihood gain; the resulting mask restricts clean-to-corrupt-prefix KL matching while CLM remains active throughout.}
    \label{fig:framework}
\end{figure*}

The key requirement is selectivity. An anti-interference objective should stabilize predictions under nuisance changes when the local suffix is sufficient. \method{} implements this with a full--short likelihood gate that separates locally sufficient targets from positions with far-prefix evidence. A fixed-reference gate analysis measures how accurately this criterion selects local-sufficient answer tokens, and prompt-level probes measure whether robustness gains concentrate in the intended regime.

The evidence follows this selectivity claim from mechanism to benchmark behavior. Mechanism analyses show that the fixed-reference gate identifies locally sufficient targets and that \method\ reduces corrupt-prefix sensitivity where this gate applies. Continued training improves \ruler\ and HELMET across five pretrained base models, and pretraining experiments improve both \ruler\ and \nolima\ across 0.3B--1B scales. Together, these results support the main claim: robust context use benefits from an explicit training signal that separates nuisance prefixes from decisive distant evidence.

Our contributions are:
\begin{itemize}
    \item We formulate far-prefix interference as selective conditional dependence: locally sufficient predictions should resist nuisance prefixes, and the full--short gap identifies where this invariance signal applies.
    \item We propose \method, a pretraining objective that corrupts only the far prefix, preserves the local suffix, and applies gated distribution matching to locally sufficient suffix tokens.
    \item We show that the learned behavior is selective: the gate identifies locally sufficient targets, and prefix sensitivity decreases sharply on the selected region.
    \item We validate \method\ across both from-scratch pretraining and continued training on pretrained base models, improving \ruler\ across five pretrained base models and improving both \ruler\ and \nolima\ in the pretraining study.
\end{itemize}

\begin{table}[t]
\centering
\captionsetup{font=footnotesize,skip=2pt}
\scriptsize
\renewcommand{\arraystretch}{0.92}
\setlength{\tabcolsep}{3pt}
\setlength{\aboverulesep}{0.25pt}
\setlength{\belowrulesep}{0.25pt}
\resizebox{\columnwidth}{!}{%
\begin{tabular}{@{}llrrrrrcc@{}}
\toprule
\multirow{2}{*}{Scale}
& \multirow{2}{*}{Benchmark}
& \multicolumn{5}{c}{$\Delta$ by context}
& \multicolumn{2}{c}{Avg score} \\
\cmidrule(lr){3-7}\cmidrule(l){8-9}
& & 1K & 2K & 3K & 4K & Avg & CE & SPAR \\
\midrule
\multirow{2}{*}{0.3B}
& \ruler{}  & +3.61 & +3.87 & +3.69 & +3.33 & +3.62 & 18.59 & 22.21 \\
& \nolima{} & +2.32 & +2.94 & +1.69 & +0.98 & +1.98 &  5.45 &  7.44 \\
\midrule
\multirow{2}{*}{0.6B}
& \ruler{}  & +1.97 & +0.74 & +1.37 & +0.85 & +1.23 & 28.54 & 29.78 \\
& \nolima{} & +4.05 & +3.05 & +1.17 & +1.71 & +2.50 &  5.37 &  7.87 \\
\midrule
\multirow{2}{*}{1B}
& \ruler{}  & +3.51 & +3.96 & +2.97 & +2.41 & +3.21 & 36.34 & 39.56 \\
& \nolima{} & +5.34 & +2.00 & +1.46 & +5.58 & +3.60 &  9.40 & 13.00 \\
\bottomrule
\end{tabular}}
\caption{Pretraining results averaged over three seeds. Context columns report gains (SPAR minus CE); final columns give absolute average scores. CE denotes ordinary CLM training.}
\label{tab:pretrain-ruler-nolima}
\end{table}

\begin{table}[t]
\centering
\captionsetup{font=footnotesize,skip=2pt}
\scriptsize
\renewcommand{\arraystretch}{0.92}
\setlength{\tabcolsep}{3pt}
\setlength{\aboverulesep}{0.25pt}
\setlength{\belowrulesep}{0.25pt}
\resizebox{\columnwidth}{!}{%
\begin{tabular}{@{}lrrrrrrrcc@{}}
\toprule
\multirow{2}{*}{Model}
& \multicolumn{7}{c}{$\Delta$ by context}
& \multicolumn{2}{c}{Avg score} \\
\cmidrule(lr){2-8}\cmidrule(l){9-10}
& 4K & 8K & 12K & 16K & 24K & 32K & Avg & CE & SPAR \\
\midrule
Qwen2.5-0.5B & +0.24 & +3.32 & +1.48 & +3.55 & -- & -- & +2.15 & 57.83 & 59.98 \\
Qwen2.5-3B   & +0.37 & +0.14 & +0.07 & +0.45 & +0.38 & +0.55 & +0.33 & 79.13 & 79.46 \\
Llama-3.2-1B & +1.20 & +3.78 & +4.98 & +4.01 & +3.54 & +2.75 & +3.38 & 64.16 & 67.53 \\
Llama-3.1-8B & +0.62 & +1.84 & +2.46 & +2.10 & +1.72 & +1.38 & +1.69 & 78.42 & 80.11 \\
GPT2-XL      & +2.77 & +2.03 & +3.06 & +3.61 & +3.53 & +5.59 & +3.43 & 30.87 & 34.30 \\
\bottomrule
\end{tabular}}
\caption{Continued training on pretrained base models averaged over three seeds. Context columns report \ruler{} gains (SPAR minus CE); final columns give absolute average scores. CE denotes ordinary CLM training.}
\label{tab:main-cpt-ruler}
\end{table}

\section{Related Work}

\paragraph{Making longer context computable.}
A large part of long-context research expands the sequence length that a Transformer can process. Early work addressed the fixed-window limitation through recurrence or sparse attention, as in Transformer-XL, Reformer, Longformer, and BigBird \citep{dai2019transformerxl,kitaev2020reformer,beltagy2020longformer,zaheer2020bigbird}. Systems and kernel work, represented by FlashAttention, made exact attention substantially more practical at longer lengths \citep{dao2022flashattention}. Another line extends or modifies positional behavior: RoPE became a standard positional mechanism for decoder-only LMs \citep{su2021roformer}; ALiBi enables input-length extrapolation through linear attention biases \citep{press2022alibi}; Position Interpolation, YaRN, LongRoPE, and PoSE adapt position handling so pretrained models can operate beyond their original windows \citep{chen2023position,peng2023yarn,ding2024longrope,zhu2023pose}. Efficient adaptation and deployment methods such as LongLoRA and StreamingLLM further reduce the cost of long-context use \citep{chen2023longlora,xiao2024streamingllm}. Alignment and data recipes including LongAlign, SkipAlign, LongWriter, and LongCite show that context behavior also depends on the length patterns and supervision formats seen during instruction tuning or continued training \citep{bai2024longalign,wu2024skipalign,bai2024longwriter,zhang2024longcite}. This literature establishes the infrastructure for longer inputs. \method\ adds an objective-level training signal for deciding when visible distant tokens should leave a locally supported prediction stable.

\paragraph{Long-context benchmarks reveal selective context-use challenges.}
Benchmark work has steadily sharpened the distinction between nominal context length and effective context use. SCROLLS and ZeroSCROLLS moved evaluation toward naturally long texts and zero-shot long-text understanding \citep{shaham2022scrolls,shaham2023zeroscrolls}. LongBench broadened long-context evaluation across bilingual, multitask settings \citep{bai2023longbench}. LooGLE and BAMBOO emphasized long-dependency QA, hallucination detection, text sorting, language modeling, and code completion over newer and longer documents \citep{li2023loogle,dong2023bamboo}. InfiniteBench pushed evaluation beyond 100K-token contexts \citep{zhang2024infinitebench}. NeedleBench, LongICLBench, Loong, and LongBench v2 further stress retrieval, reasoning chains, extreme-label in-context learning, extended multi-document QA, and deeper reasoning \citep{li2024needlebench,li2024longiclbench,wang2024loong,bai2024longbenchv2}. RULER provides structured 13-task stress tests that expose degradation across retrieval, multi-hop, aggregation, and QA-like settings \citep{hsieh2024ruler}. HELMET argues for holistic long-context evaluation because synthetic tasks and application tasks can produce different rankings \citep{yen2024helmet}. NoLiMa removes literal lexical matching between question and evidence, making the benchmark closer to association-based retrieval \citep{modarressi2025nolima}. Lost-in-the-middle results show that models use long inputs unevenly even when the answer is present \citep{liu2024lost}. Taken together, these benchmarks show a recurring pattern: models can accept more tokens while remaining fragile about which distant evidence they use, where that evidence appears, and how much irrelevant material surrounds it.

\paragraph{Training objectives for useful context dependence.}
Several recent methods improve behavior in long, packed, or distractor-heavy contexts by changing the training signal. LongAlign constructs long instruction-following data and batching strategies for long-context alignment \citep{bai2024longalign}; SkipAlign synthesizes position gaps to expose models to long-range dependencies without full-length training examples \citep{wu2024skipalign}; LongWriter and LongCite show that output length and citation faithfulness depend strongly on the supervision distribution \citep{bai2024longwriter,zhang2024longcite}. LongPPL and LongCE are especially close to the objective-level question. They argue that standard perplexity averages away the tokens that actually benefit from long context, then use long-short contrast to identify and upweight key tokens \citep{fang2024longppl}. \method{} uses the long--short signal to identify locally sufficient tokens whose predictions should remain stable under irrelevant far-prefix changes, turning the contrast into an anti-interference target.

\paragraph{Irrelevant context, distractibility, and far-prefix interference.}
The motivation for \method\ is also connected to work on distractors and irrelevant context. Large language models can be distracted by irrelevant information in reasoning problems \citep{shi2023distracted}. Position sensitivity in long inputs \citep{liu2024lost}, association-based retrieval in NoLiMa \citep{modarressi2025nolima}, and benchmark disagreements documented by HELMET \citep{yen2024helmet} all point to the same underlying issue: context utilization is a selection problem. A model must preserve useful long-range dependencies while preventing irrelevant spans from steering predictions that are already locally determined. Existing benchmarks diagnose this behavior at evaluation time. \method\ turns it into a pretraining-time objective by constructing clean and far-prefix-corrupted views and optimizing local-suffix predictions to remain stable only when the gate indicates local sufficiency.

\paragraph{Consistency and invariance training.}
\method\ is related in form to consistency regularization, where models are trained to produce stable predictions under perturbations. Virtual Adversarial Training enforces smoothness around inputs \citep{miyato2018vat}; Mean Teacher uses weight-averaged teachers to form consistency targets \citep{tarvainen2017mean}; UDA improves semi-supervised learning by enforcing invariance under strong data augmentations \citep{xie2020uda}; R-Drop regularizes dropout-induced subnetworks through bidirectional KL matching \citep{liang2021rdrop}. \method\ adapts this idea to a structured far-prefix threat model. The perturbation changes the far prefix while preserving the local suffix, and the target is the subset selected by a short-context sufficiency gate. This selectivity ties the consistency loss to anti-interference in long, packed, or distractor-heavy contexts.

\begin{table}[t]
\centering
\begingroup
\captionsetup{font=footnotesize,skip=2pt}
\tiny
\renewcommand{\arraystretch}{0.92}
\setlength{\tabcolsep}{3pt}
\setlength{\aboverulesep}{0.25pt}
\setlength{\belowrulesep}{0.25pt}
\resizebox{0.94\columnwidth}{!}{%
\begin{tabular}{@{}lrrrrcc@{}}
\toprule
\multirow{2}{*}{Model}
& \multicolumn{4}{c}{$\Delta$ by context}
& \multicolumn{2}{c}{Avg score} \\
\cmidrule(lr){2-5}\cmidrule(l){6-7}
& 8K & 16K & 32K & Avg & CE & SPAR \\
\midrule
Qwen2.5-0.5B & +1.49 & +1.84 & +1.46 & +1.60 & 33.83 & 35.43 \\
Qwen2.5-3B   & +0.71 & +0.55 & +0.32 & +0.53 & 59.98 & 60.51 \\
Llama-3.2-1B & +1.03 & +2.21 & +1.55 & +1.60 & 50.53 & 52.13 \\
Llama-3.1-8B & +0.48 & +1.22 & +0.86 & +0.85 & 63.74 & 64.59 \\
GPT2-XL   & +1.07 & +2.49 & +1.31 & +1.62 & 16.43 & 18.06 \\
\bottomrule
\end{tabular}}
\caption{HELMET long-context evaluation averaged over three seeds on NQ, TriviaQA, HotpotQA, PopQA, NIAH multikey-2/3, NIAH multivalue, JSON-KV, and MS MARCO reranking. Context columns report macro gains (SPAR minus CE); final columns give absolute average scores. CE denotes ordinary CLM training.}
\label{tab:helmet-cpt}
\endgroup
\end{table}

\begin{table*}[t]
\centering
\tiny
\renewcommand{\arraystretch}{0.92}
\setlength{\tabcolsep}{2.6pt}
\resizebox{0.88\textwidth}{!}{%
\begin{tabular}{llrrrrrrrrrrr}
\toprule
Model & SPAR & PIQA & SIQA & Hella & Wino & ARC-e & ARC-c & OBQA & RACE & Lambada & MMLU & Avg \\
\midrule
\multirow{2}{*}{0.3B}
& $\times$ & 56.94 & 42.51 & 26.18 & 49.12 & 28.51 & 21.07 & 27.92 & 25.84 & 3.92 & 20.42 & 30.24 \\
& $\checkmark$ & 63.72 & 40.62 & 28.54 & 47.92 & 36.86 & 22.41 & 31.08 & 28.48 & 13.92 & 20.11 & \textbf{33.37} \\
\midrule
\multirow{2}{*}{0.6B}
& $\times$ & 66.52 & 43.18 & 32.94 & 49.74 & 39.82 & 25.08 & 30.54 & 30.88 & 13.96 & 22.73 & 35.54 \\
& $\checkmark$ & 69.27 & 41.73 & 35.29 & 49.49 & 46.91 & 26.42 & 32.79 & 31.33 & 20.36 & 23.91 & \textbf{37.75} \\
\midrule
\multirow{2}{*}{1B}
& $\times$ & 73.18 & 44.12 & 39.62 & 50.08 & 47.58 & 28.76 & 32.52 & 34.18 & 23.82 & 24.78 & 39.87 \\
& $\checkmark$ & 73.52 & 42.18 & 40.52 & 50.58 & 54.82 & 29.77 & 32.86 & 32.22 & 25.14 & 26.08 & \textbf{40.77} \\
\midrule
\rowcolor{blue!12}
Average $\Delta$Acc
& -- & +3.29 & -1.76 & +1.87 & -0.32 & +7.56 & +1.23 & +1.92 & +0.38 & +5.90 & +0.72 & \textbf{+2.08} \\
\bottomrule
\end{tabular}%
}
\caption{Zero-shot public benchmark accuracy (\%) for from-scratch pretrained decoder-only Transformer base models.}
\label{tab:public-mc-base}
\end{table*}

\section{Selective Prefix Anti-Interference Regularization}
\label{sec:method}

\subsection{Far-Prefix Interference and Conditional Dependence}

Let $x=(x_1,\ldots,x_T)$ be a training sequence. We split it into a far prefix $x_{1:a}$ and a local suffix $x_{a+1:T}$. For a suffix position $t>a$, a decoder-only language model predicts $x_t$ from $x_{<t}$. Far-prefix interference occurs when modifying $x_{1:a}$ substantially changes the prediction for $x_t$ even though the recent local context already supports that prediction.

This definition isolates prefix-induced instability from irrelevant distant context and gives a direct training target. Let $F=x_{1:a}$ denote the far prefix, $L=x_{a+1:t-1}$ the local suffix, and $Y=x_t$ the target. For the observed target token, the model-based quantity
\begin{equation}
    \iota_t(Y;F\mid L)
    = \log p_\theta(Y\mid F,L)-\log p_\theta(Y\mid L)
\end{equation}
is a pointwise estimate of the additional predictive information supplied by $F$; its expectation parallels conditional mutual information. Large $\iota_t$ indicates predictive evidence in the far prefix. Small $\iota_t$ together with high local predictability identifies an invariance region in which nuisance changes to $F$ should leave the prediction stable. \method\ targets this low-conditional-information region.

\subsection{Clean and Corrupt-Prefix Views}

\method\ constructs a corrupt-prefix input for a subset of training examples. The clean input is the original sequence $x$. The corrupt-prefix input is
\begin{equation}
    \tilde{x} = [c(x_{1:a}), x_{a+1:T}],
\end{equation}
where $c(\cdot)$ corrupts the far prefix by perturbing its chunk order while preserving the local suffix. The clean and corrupt-prefix inputs therefore predict the same suffix tokens.

This construction fixes the target suffix and its immediate evidence while changing only the distant prefix. Because both views predict identical suffix tokens, the KL term compares two distributions over the same next-token labels and recent suffix evidence. The paired comparison converts far-prefix interference into a direct training signal at locally sufficient positions.

The model computes clean logits $z_t$ and corrupt-prefix logits $\tilde{z}_t$ for suffix positions. The ordinary CLM loss remains the primary training objective:
\begin{equation}
    \clm = - \sum_t \log p_\theta(x_t \mid x_{<t}).
\end{equation}

\subsection{Short-Context Sufficiency Gate}

The invariance loss is applied on positions where local context is sufficient. For each suffix token $x_t$, we evaluate a short-context view containing the most recent $K$ tokens before $t$. Let $p_{\mathrm{short}}(x_t)$ be the gold-token probability under this short view, and let $p_{\mathrm{full}}(x_t)$ be the gold-token probability under the full clean context. The gate is
\begin{equation}
    \begin{aligned}
    g_t = \mathbf{1}[&
        p_{\mathrm{short}}(x_t) \ge \tau_{\mathrm{abs}}
        \;\wedge \\
        &\log p_{\mathrm{full}}(x_t)
        - \log p_{\mathrm{short}}(x_t)
        \le \tau_{\mathrm{gap}}].
    \end{aligned}
\end{equation}
The first condition selects locally predictable tokens. The second condition filters positions where the full context provides substantial additional evidence. The gate is the key selectivity mechanism: tokens that genuinely depend on long-range context remain governed by ordinary CLM.

The two gate conditions play different roles. High short-context likelihood prevents the objective from regularizing uncertain suffix positions, where instability may reflect weak local evidence. The full--short gap then separates local sufficiency from genuine long-range dependence: a token can be predictable under the short view while still gaining decisive evidence from the far prefix. Applying invariance only when both conditions hold makes the loss asymmetric with respect to context use. It suppresses nuisance dependence where the local suffix already determines the target, while leaving high-gap positions to ordinary CLM.

\subsection{Gated Invariance Loss}

\method\ matches clean and corrupt-prefix distributions on gated positions:
\begin{equation}
    \linv =
    \frac{\sum_{t>a} g_t \,
    \kl\left(
        \mathrm{sg}(p_\theta(\cdot \mid x_{<t}))
        \,\|\, p_\theta(\cdot \mid \tilde{x}_{<t})
    \right)}
    {\max(\sum_{t>a} g_t, 1)}.
\end{equation}
The invariance term uses a directional anchor. The clean-view distribution is evaluated with stop-gradient and serves as the reference distribution; the corrupt-prefix branch receives the consistency gradient. This directs the auxiliary update toward removing prefix-induced drift while keeping the ordinary clean-context CLM loss as the source of token likelihood learning. Since the two branches share parameters, gradients through the corrupt branch still update the same model used at clean-context inference time. For top-$k$ matching, the support is selected from the clean anchor, both distributions are renormalized over this support, and the anchor temperature controls the sharpness of the reference before truncation.

The final objective is
\begin{equation}
    \mathcal{L} = \clm + \lambda_{\mathrm{inv}}\linv.
\end{equation}
Experiments use top-$k=32$ and anchor temperature $0.7$, with corrupt-prefix CLM disabled. The SPAR batch fraction controls the fraction of batches that carry the additional corrupt-prefix computation.

\subsection{Evaluating Selective Context Dependence}

The prefix-sensitivity analysis uses the same clean/corrupt-prefix geometry. For validation sequences, we compute gold-token log likelihood under the clean view and under the corrupt-prefix view. The sensitivity drop is
\begin{equation}
    \Delta_{\mathrm{sens}}
    =
    \mathbb{E}_{t}
    \left[
        \log p_\theta(x_t \mid x_{<t})
        -
        \log p_\theta(x_t \mid \tilde{x}_{<t})
    \right].
\end{equation}
Lower sensitivity drop means far-prefix corruption harms the gold-token probability less. The gate-selected version restricts the same measurement to positions selected by the local-sufficiency criterion:
\begin{equation}
    \Delta_{\mathrm{gated}}
    =
    \frac{
    \sum_t g_t
    \left[
        \log p_\theta(x_t \mid x_{<t})
        -
        \log p_\theta(x_t \mid \tilde{x}_{<t})
    \right]}
    {\max(\sum_t g_t, 1)}.
\end{equation}
We report both all-token sensitivity and gated sensitivity. Gated sensitivity is the closer mechanism check because it measures the positions selected by the same local-sufficiency criterion used in training. To evaluate the gate without model-dependent subset shift, the gate-region analysis computes token bins and the evaluation mask once with a frozen CE reference model, then applies the same tokens, bins, and mask to CE and \method.

\section{Experimental Setup}

\paragraph{Continued training on pretrained base models.}
The main experiments compare \method\ against a CE baseline (ordinary CLM training) for continued training on pretrained base models with equal counted training compute. We evaluate Qwen2.5-0.5B, Qwen2.5-3B, Llama-3.2-1B, Llama-3.1-8B, and GPT2-XL. Qwen and Llama models use a 16K training window, and GPT2-XL uses a 32K training window. The main \ruler\ evaluation is run directly on the resulting base models.

\paragraph{Pretraining.}
We evaluate 0.3B, 0.6B, and 1B pretraining experiments on both \ruler\ and \nolima. These experiments test the same anti-interference objective before continued training on pretrained base models.

\paragraph{Benchmarks and metrics.}
\ruler\ is the primary benchmark for continued training on pretrained base models. It contains 13 long-context tasks spanning retrieval, multi-hop tracing, aggregation, and QA-style synthetic evaluation \citep{hsieh2024ruler}. We report task macro averages at each context length and macro deltas between \method\ and the corresponding baseline. HELMET provides a secondary long-context evaluation on application-oriented QA, retrieval, key-value lookup, and reranking tasks at 8K--32K \citep{yen2024helmet}. \nolima\ is used in the pretraining experiments to test retrieval beyond literal matching \citep{modarressi2025nolima}. The prefix-sensitivity analysis measures the gold-token likelihood drop caused by far-prefix corruption.

\paragraph{Mechanism analyses.}
We compare CE and \method\ at a 4K context length, with 3K tokens assigned to the far prefix. The per-prompt selectivity analysis uses 256 entity--value examples. Its local-sufficient condition changes nuisance prefix content while preserving a decisive local update; its far-dependent condition switches the only answer-bearing evidence in the far prefix. All candidates are scored by sequence log probability.

The gate-region analysis uses 256 held-out pretraining sequences and averages document replacement and chunk-shuffle corruption. Short NLL and the full--short log-probability gap are computed with the frozen CE model; this reference also defines a single evaluation gate shared by both target models. The fixed mask keeps CE and \method{} on the same selected token subset.

\paragraph{Evaluation windows.}
For Qwen and Llama models trained to 16K, evaluations at 24K and 32K are extrapolation measurements. For GPT2-XL, all reported \ruler\ lengths use the 32K-trained model. The main trained-window claim uses lengths up to the training window, while full reported deltas include all completed lengths when stated.

\begin{table}[t]
\centering
\begingroup
\captionsetup{font=footnotesize,skip=2pt}
\footnotesize
\setlength{\tabcolsep}{4.0pt}
\resizebox{\columnwidth}{!}{%
\begin{tabular}{lrrrrr}
\toprule
SPAR batch fraction & 4K & 8K & 12K & 16K & Macro \\
\midrule
10\% & +0.0354 & +1.6123 & +0.5862 & +2.7177 & +1.2379 \\
30\% & -0.3477 & +2.1315 & +0.8277 & +2.4985 & +1.2775 \\
50\% & +0.2608 & +3.2685 & +1.1185 & +3.8900 & +2.1344 \\
70\% & +0.2446 & +3.3231 & +1.4777 & +3.5515 & +2.1492 \\
90\% & +0.2785 & +3.5731 & +1.4754 & +3.0069 & +2.0835 \\
\bottomrule
\end{tabular}%
}
\caption{Qwen2.5-0.5B SPAR batch fraction sensitivity at fixed $\lambda_{\mathrm{inv}}=0.004$. Values are \ruler{} macro deltas against the corresponding CE baseline at each context length. CE denotes ordinary CLM training.}
\label{tab:qwen05-ratio}
\endgroup
\end{table}

\begin{figure}
    \centering
    \includegraphics[width=\linewidth]{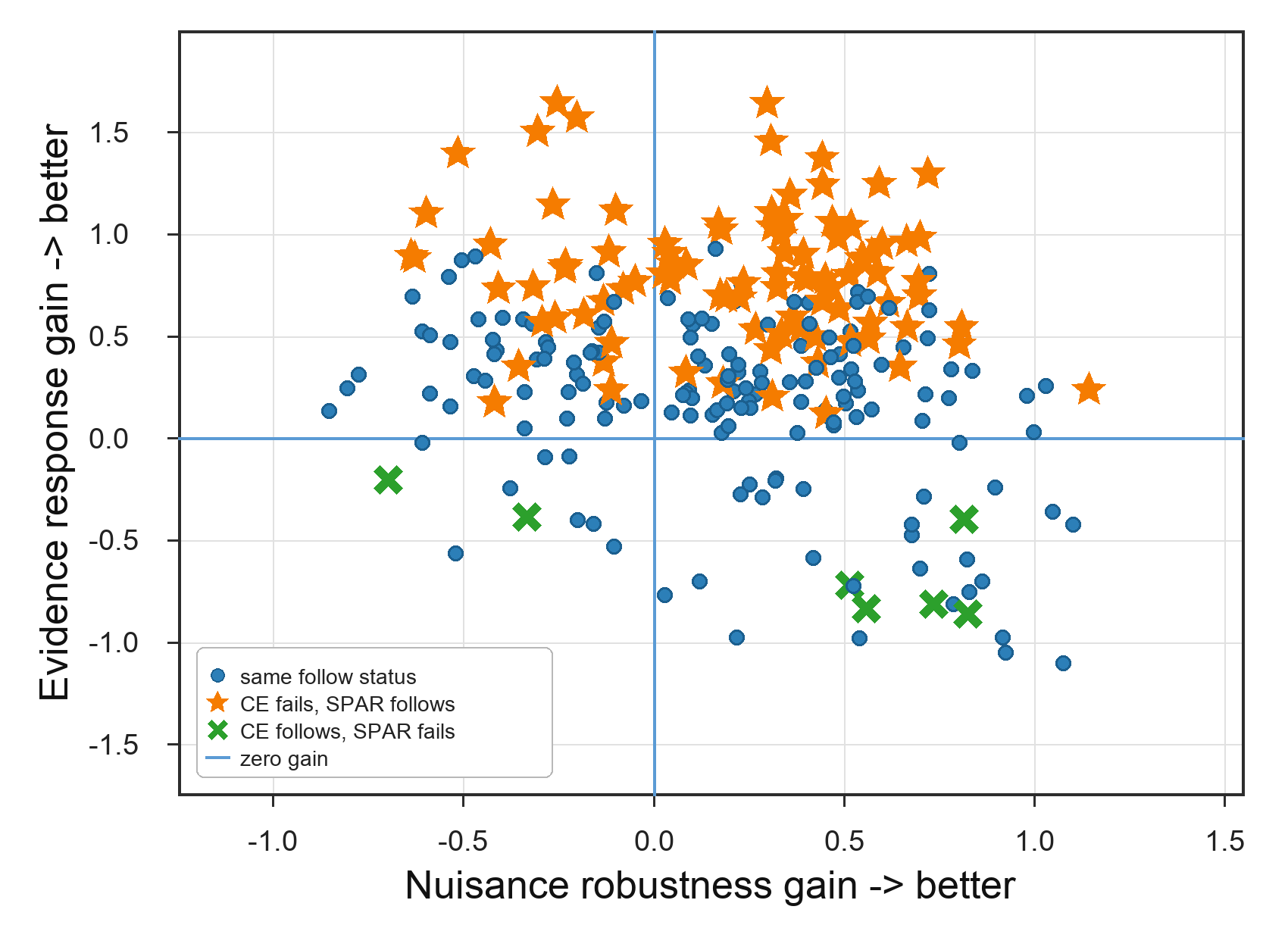}
    \caption{Per-prompt selectivity gains. Each point is one entity--value prompt. The horizontal axis measures robustness gain under nuisance prefix changes; the vertical axis measures gain in following changed answer evidence. The upper-right region marks prompts where \method{} improves both nuisance invariance and evidence following.}
    \label{fig:prompt-gain-map}
\end{figure}

\begin{figure*}[t!]
    \centering
    \includegraphics[width=0.96\textwidth]{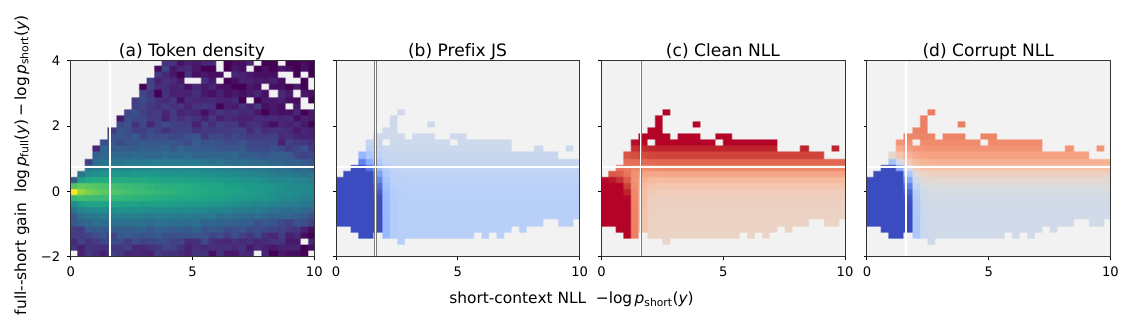}
    \caption{Gate-region analysis using an evaluation mask fixed by the frozen CE reference. Panels show (a) validation-token density, (b) \method{} minus CE prefix-distribution JS, (c) clean NLL, and (d) corrupt NLL. Blue denotes a reduction. Threshold lines mark the short-NLL and full--short-gap cutoffs. On the balanced answer-token set, the gate reaches 0.988 precision, 0.613 recall, 0.008 far-dependent false-positive rate, and 0.933 AUROC.}
    \label{fig:gate-phase}
\end{figure*}

\section{Results}

The results test whether an anti-interference objective improves long-context behavior through a selective training signal. We first evaluate the external effect in from-scratch pretraining and continued training on pretrained base models. We then localize the effect at the token and prompt levels, followed by sensitivity sweeps that characterize the useful strength range of the invariance signal.

\subsection{Pretraining Improves RULER and NoLiMa}

Table~\ref{tab:pretrain-ruler-nolima} reports pretraining gains by context length and average absolute scores across 0.3B, 0.6B, and 1B scales. \method\ improves both \ruler\ and \nolima\ at every scale, covering structured long-context retrieval as well as evidence selection without direct lexical matching.

\subsection{Continued Training Improves RULER Across Pretrained Base Models}

Table~\ref{tab:main-cpt-ruler} reports \ruler\ gains by context length and average absolute scores for continued training on pretrained base models. \method\ improves \ruler\ in every family. Table~\ref{tab:helmet-cpt} extends the comparison to HELMET at 8K--32K, where average gains remain positive for all five base models. The gains are small but consistent for Qwen2.5-3B, larger for Qwen2.5-0.5B and Llama-3.2-1B, positive but smaller on Llama-3.1-8B, and strongest for GPT2-XL at the 32K window. The shared pattern across Qwen2.5, Llama, and GPT-2 XL links from-scratch pretraining and continued training on pretrained base models to the same selective anti-interference signal.

\begin{table}[t]
\centering
\captionsetup{font=footnotesize,skip=2pt}
\scriptsize
\setlength{\tabcolsep}{4.2pt}
\renewcommand{\arraystretch}{0.95}
\begin{tabular}{lccc}
\toprule
Model & \method{} gate & Random gate & Ungated KL \\
\midrule
Qwen2.5-0.5B & +2.15 & +0.98 & +0.48 \\
Qwen2.5-3B & +0.33 & +0.12 & +0.03 \\
Llama-3.2-1B & +3.38 & +1.56 & +0.75 \\
Llama-3.1-8B & +1.69 & +0.74 & +0.34 \\
GPT2-XL & +3.43 & +1.55 & +0.85 \\
\bottomrule
\end{tabular}
\caption{\ruler{} ablation of the selection rule. Values are average gains over CE for each model, using the same evaluated context lengths as Table~\ref{tab:main-cpt-ruler}. CE denotes ordinary CLM training.}
\label{tab:gate-selection-ablation}
\end{table}

\subsection{Gate Selection Matters}

Table~\ref{tab:gate-selection-ablation} separates the value of the gate from the value of adding an auxiliary consistency loss. Random gating keeps the same selection budget but assigns positions independently of local sufficiency; ungated KL keeps clean--corrupt matching but applies it to every suffix position. The \method{} gate selects targets by the short-context likelihood and the full--short likelihood gap, so the agreement loss is concentrated where the suffix already contains enough evidence for prediction and avoids positions whose prediction can legitimately depend on the far prefix.

This selection rule gives the largest average \ruler{} gain for every model. The ordering is strongest on Qwen2.5-0.5B, Llama-3.2-1B, and GPT2-XL: \method{} reaches +2.15, +3.38, and +3.43 average gain, compared with +0.98, +1.56, and +1.55 for random gating. Ungated KL is weaker across the same models. The consistent ordering supports the main training mechanism: the useful anti-interference signal comes from matching clean and corrupt-prefix predictions at locally sufficient suffix positions.

\subsection{Public Benchmark Transfer}

Table~\ref{tab:public-mc-base} evaluates from-scratch pretrained decoder-only Transformer base models on public zero-shot benchmarks before instruction or QA fine-tuning. Multiple-choice tasks use log-probability scoring; Lambada uses final-word exact accuracy. \method{} improves the average score at all three scales, with gains concentrated on PIQA, HellaSwag, ARC, OpenBookQA, Lambada, and MMLU.

\subsection{Evidence for Selective Context Dependence}

The token- and prompt-level analyses localize the training effect behind the benchmark gains. Gate-region measurements identify the answer-token region selected by the objective, and prompt-level probes measure whether robustness improves in the intended local-sufficiency regime.

\paragraph{Where the objective acts.}
The frozen-reference gate is conservative and concentrates on local-sufficient answer positions (Figure~\ref{fig:gate-phase}). Table~\ref{tab:gate-region-summary} aggregates the same mask into selected and unselected regions.

\begin{table}[t]
\centering
\scriptsize
\setlength{\tabcolsep}{2.3pt}
\renewcommand{\arraystretch}{0.96}
\begin{tabular}{@{}lrrrr@{}}
\toprule
Region & $\Delta$ JS & $\Delta$ Clean & $\Delta$ Corrupt & $\Delta$ Sens. \\
\midrule
Gate-selected & -0.0049 & +0.0186 & -0.0634 & -0.0820 \\
Unselected & -0.0005 & +0.0021 & -0.0019 & -0.0040 \\
All suffix & -0.0014 & +0.0048 & -0.0096 & -0.0144 \\
\bottomrule
\end{tabular}
\caption{Gate-region aggregate for Figure~\ref{fig:gate-phase}. Values are \method{} minus CE under the fixed-reference mask; $\Delta$ Sens. is the change in corrupt--clean NLL gap. CE denotes ordinary CLM training.}
\label{tab:gate-region-summary}
\end{table}

The selected region shows a much larger per-token reduction than the unselected region: prefix JS falls by 0.0049 versus 0.0005, and sensitivity falls by 0.0820 versus 0.0040; the small clean-view NLL increase (+0.0186) is paired with a larger corrupt-view NLL decrease (-0.0634), matching the directional anchor in the invariance loss. Because this mask is fixed by the CE reference before comparing models, CE and \method{} are evaluated on the same selected tokens.

Figure~\ref{fig:prompt-gain-map} lifts the same question from tokens to prompts. Points in the right half indicate prompts where \method{} is less sensitive to irrelevant prefix changes than CE; points in the upper half indicate prompts where it better follows a changed answer-bearing fact. Gains concentrate in the upper-right region, and evidence-switch fixes substantially outnumber losses. The prompt-level pattern matches the gate-region evidence: the training signal improves nuisance invariance while retaining responsiveness to answer-changing evidence.

\subsection{SPAR Strength Sensitivity}

Qwen2.5-0.5B varies the fraction of batches that carry \method{} supervision at fixed $\lambda_{\mathrm{inv}}=0.004$. Table~\ref{tab:qwen05-ratio} shows small gains at 10\%--30\%, followed by a broad high-gain region at 50\%--90\%. The largest gains appear at 8K--16K, matching the anti-interference motivation.

\begin{figure}
    \centering
    \includegraphics[width=0.82\linewidth]{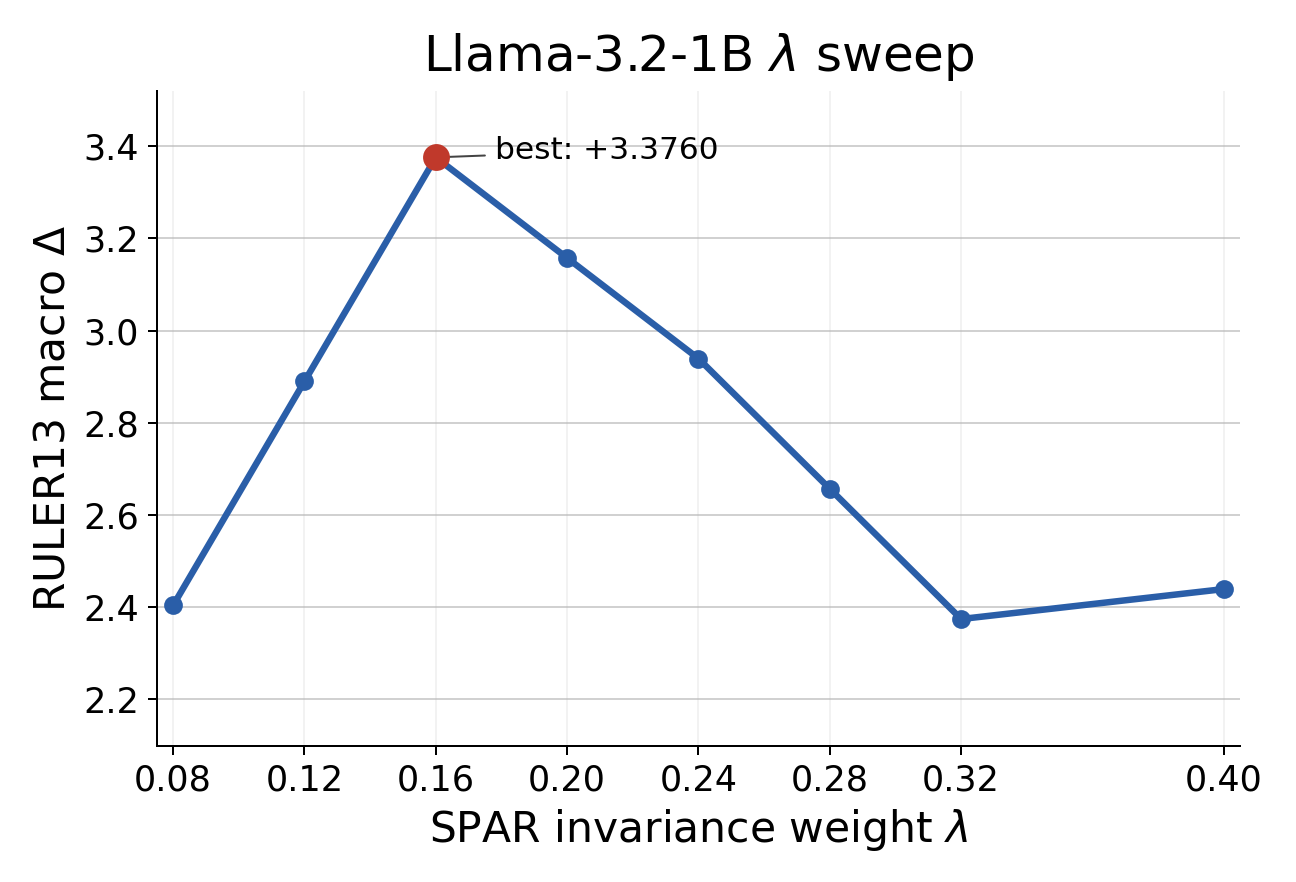}
    \caption{Llama-3.2-1B invariance-weight sweep at a fixed 10\% SPAR batch fraction. The macro \ruler\ gain peaks at moderate strength.}
    \label{fig:lambda-sweep}
\end{figure}

The auxiliary invariance signal has a useful moderate-strength range. Figure~\ref{fig:lambda-sweep} sweeps the invariance weight on Llama-3.2-1B while holding the SPAR batch fraction fixed at 10\%. The gain rises as $\lambda_{\mathrm{inv}}$ increases from 0.008 to 0.016, remains high near 0.020, and then declines at larger weights. The peak marks the useful strength range for the selective regularizer.

\section{Discussion}

\paragraph{Selective anti-interference.}
\method\ adds an explicit anti-interference bias to pretraining: predictions that are locally supported should stay stable when irrelevant distant text changes. The full--short log-probability gap gives this principle a conditional-dependence interpretation. Low-gap, locally predictable targets define the region in which invariance is desirable; high-gap targets identify positions where the far prefix supplies predictive evidence. The mechanism analyses connect this token-level criterion to behavior: the gate identifies local sufficiency, and corrupt-prefix sensitivity drops in the selected region.

\paragraph{Why the gate matters.}
The local-sufficiency gate makes the objective selective. Its 0.988 precision and 0.008 far-dependent false-positive rate on the balanced answer-token set indicate conservative selection. The mechanism analyses show fewer target flips in the local-sufficient regime, stronger selected-region stability, and a broad reduction in gate-targeted prefix sensitivity. Together, these results support the intended behavior: lower irrelevant-prefix dependence where the local suffix already supports the prediction.

\paragraph{Relation to key-token objectives.}
Key-token objectives such as LongCE emphasize positions whose likelihood improves from long context. \method\ acts on the complementary region identified by the full--short gap: locally sufficient targets where stable prediction under far-prefix changes is the desired behavior. This places anti-interference training on a distinct axis from long-context token upweighting.

\paragraph{Objective-level view.}
CLM supplies the pressure to exploit predictive context. \method{} adds a selective pressure to damp irrelevant far-prefix variation when the suffix already carries enough evidence. The resulting training signal is local in where it acts and global in what it discourages: it does not suppress long-range dependence uniformly, but only reduces sensitivity to far-prefix changes at positions whose short-context evidence is already sufficient. This distinction matters for long-context learning because useful evidence and nuisance text often occupy the same sequence. This objective-level signal explains the selected-region sensitivity drop and the gains across pretraining and continued training on pretrained base models.

\section{Conclusion}

We presented Selective Prefix Anti-Interference Regularization, a selective anti-interference objective for causal LM pretraining. A short-context gate identifies locally sufficient targets and stabilizes them under far-prefix corruption. Mechanism analyses show conservative gate selection and lower selected-region prefix sensitivity; benchmark results show gains across continued training on pretrained base models and from-scratch pretraining. Selective anti-interference pretraining provides a direct objective-level signal for robust context use.

Robust context use depends on what a model learns to ignore as well as what it learns to retrieve. \method{} adds this complementary signal without changing the model architecture or requiring task-specific supervision.

\bibliography{references}

@article{liu2024lost,
  title={Lost in the middle: How language models use long contexts},
  author={Liu, Nelson F and Lin, Kevin and Hewitt, John and Paranjape, Ashwin and Bevilacqua, Michele and Petroni, Fabio and Liang, Percy},
  journal={arXiv preprint arXiv:2307.03172},
  year={2023}
}

@article{hsieh2024ruler,
  title={RULER: What's the real context size of your long-context language models?},
  author={Hsieh, Cheng-Ping and Sun, Simeng and Kriman, Samuel and Acharya, Shantanu and Rekesh, Dima and Jia, Fei and Zhang, Yang and Ginsburg, Boris},
  journal={arXiv preprint arXiv:2404.06654},
  year={2024}
}

@article{bai2023longbench,
  title={Longbench: A bilingual, multitask benchmark for long context understanding},
  author={Bai, Yushi and Lv, Xin and Zhang, Jiajie and Lyu, Hongchang and Tang, Jiankai and Huang, Zhidian and Du, Zhengxiao and Liu, Xiao and Zeng, Aohan and Hou, Lei and others},
  journal={arXiv preprint arXiv:2308.14508},
  year={2023}
}

@article{modarressi2025nolima,
  title={Nolima: Long-context evaluation beyond literal matching},
  author={Modarressi, Ali and Deilamsalehy, Hanieh and Dernoncourt, Franck and Bui, Trung and Rossi, Ryan A and Yoon, Seunghyun and Sch{\"u}tze, Hinrich},
  journal={arXiv preprint arXiv:2502.05167},
  year={2025}
}

@article{fang2024longppl,
  title={What is wrong with perplexity for long-context language modeling?},
  author={Fang, Lizhe and Wang, Yifei and Liu, Zhaoyang and Zhang, Chenheng and Jegelka, Stefanie and Gao, Jinyang and Ding, Bolin and Wang, Yisen},
  journal={arXiv preprint arXiv:2410.23771},
  year={2024}
}

@article{peng2023yarn,
  title={Yarn: Efficient context window extension of large language models},
  author={Peng, Bowen and Quesnelle, Jeffrey and Fan, Honglu and Shippole, Enrico},
  journal={arXiv preprint arXiv:2309.00071},
  year={2023}
}

@article{ding2024longrope,
  title={Longrope: Extending llm context window beyond 2 million tokens},
  author={Ding, Yiran and Zhang, Li Lyna and Zhang, Chengruidong and Xu, Yuanyuan and Shang, Ning and Xu, Jiahang and Yang, Fan and Yang, Mao},
  journal={arXiv preprint arXiv:2402.13753},
  year={2024}
}

@article{chen2023longlora,
  title={Longlora: Efficient fine-tuning of long-context large language models},
  author={Chen, Yukang and Qian, Shengju and Tang, Haotian and Lai, Xin and Liu, Zhijian and Han, Song and Jia, Jiaya},
  journal={arXiv preprint arXiv:2309.12307},
  year={2023}
}

@inproceedings{shi2023distracted,
  title={Large language models can be easily distracted by irrelevant context},
  author={Shi, Freda and Chen, Xinyun and Misra, Kanishka and Scales, Nathan and Dohan, David and Chi, Ed H and Sch{\"a}rli, Nathanael and Zhou, Denny},
  booktitle={International Conference on Machine Learning},
  pages={31210--31227},
  year={2023},
  organization={PMLR}
}

@article{xie2020uda,
  title={Unsupervised data augmentation for consistency training},
  author={Xie, Qizhe and Dai, Zihang and Hovy, Eduard and Luong, Thang and Le, Quoc},
  journal={Advances in neural information processing systems},
  volume={33},
  pages={6256--6268},
  year={2020}
}

@article{liang2021rdrop,
  title={R-drop: Regularized dropout for neural networks},
  author={Wu, Lijun and Li, Juntao and Wang, Yue and Meng, Qi and Qin, Tao and Chen, Wei and Zhang, Min and Liu, Tie-Yan and others},
  journal={Advances in neural information processing systems},
  volume={34},
  pages={10890--10905},
  year={2021}
}

@article{dai2019transformerxl,
  title={Transformer-xl: Attentive language models beyond a fixed-length context},
  author={Dai, Zihang and Yang, Zhilin and Yang, Yiming and Carbonell, Jaime and Le, Quoc V and Salakhutdinov, Ruslan},
  journal={arXiv preprint arXiv:1901.02860},
  year={2019}
}

@article{kitaev2020reformer,
  title={Reformer: The efficient transformer},
  author={Kitaev, Nikita and Kaiser, {\L}ukasz and Levskaya, Anselm},
  journal={arXiv preprint arXiv:2001.04451},
  year={2020}
}

@article{beltagy2020longformer,
  title={Longformer: The long-document transformer},
  author={Beltagy, Iz and Peters, Matthew E and Cohan, Arman},
  journal={arXiv preprint arXiv:2004.05150},
  year={2020}
}

@article{zaheer2020bigbird,
  title={Big bird: Transformers for longer sequences},
  author={Zaheer, Manzil and Guruganesh, Guru and Dubey, Kumar Avinava and Ainslie, Joshua and Alberti, Chris and Ontanon, Santiago and Pham, Philip and Ravula, Anirudh and Wang, Qifan and Yang, Li and others},
  journal={Advances in neural information processing systems},
  volume={33},
  pages={17283--17297},
  year={2020}
}

@inproceedings{dao2022flashattention,
  title={Flashattention: Fast and memory-efficient exact attention with io-awareness},
  author={Dao, Tri and Fu, Daniel Y and Ermon, Stefano and Rudra, Atri and R{\'e}, Christopher},
  booktitle={Advances in neural information processing systems},
  year={2022}
}

@article{su2021roformer,
  title={Roformer: Enhanced transformer with rotary position embedding},
  author={Su, Jianlin and Lu, Yu and Pan, Shengfeng and Murtadha, Ahmed and Wen, Bo and Liu, Yunfeng},
  journal={arXiv preprint arXiv:2104.09864},
  year={2021}
}

@article{press2022alibi,
  title={Train short, test long: Attention with linear biases enables input length extrapolation},
  author={Press, Ofir and Smith, Noah A and Lewis, Mike},
  journal={arXiv preprint arXiv:2108.12409},
  year={2021}
}

@article{chen2023position,
  title={Extending context window of large language models via positional interpolation},
  author={Chen, Shouyuan and Wong, Sherman and Chen, Liangjian and Tian, Yuandong},
  journal={arXiv preprint arXiv:2306.15595},
  year={2023}
}

@inproceedings{zhu2023pose,
  title={Pose: Efficient context window extension of llms via positional skip-wise training},
  author={Zhu, Dawei and Yang, Nan and Wang, Liang and Song, Yifan and Wu, Wenhao and Wei, Furu and Li, Sujian},
  booktitle={International Conference on Learning Representations},
  volume={2024},
  pages={18940--18954},
  year={2024}
}

@inproceedings{xiao2024streamingllm,
  title={Efficient streaming language models with attention sinks},
  author={Xiao, Guangxuan and Tian, Yuandong and Chen, Beidi and Han, Song and Lewis, Mike},
  booktitle={International Conference on Learning Representations},
  volume={2024},
  pages={21875--21895},
  year={2024}
}

@inproceedings{bai2024longalign,
  title={Longalign: A recipe for long context alignment of large language models},
  author={Bai, Yushi and Lv, Xin and Zhang, Jiajie and He, Yuze and Qi, Ji and Hou, Lei and Tang, Jie and Dong, Yuxiao and Li, Juanzi},
  booktitle={Findings of the Association for Computational Linguistics: EMNLP 2024},
  pages={1376--1395},
  year={2024}
}

@article{wu2024skipalign,
  title={Long context alignment with short instructions and synthesized positions},
  author={Wu, Wenhao and Wang, Yizhong and Fu, Yao and Yue, Xiang and Zhu, Dawei and Li, Sujian},
  journal={arXiv preprint arXiv:2405.03939},
  year={2024}
}

@inproceedings{bai2024longwriter,
  title={Longwriter: Unleashing 10,000+ word generation from long context llms},
  author={Bai, Yushi and Zhang, Jiajie and Lv, Xin and Zheng, Linzhi and Zhu, Siqi and Hou, Lei and Dong, Yuxiao and Tang, Jie and Li, Juanzi},
  booktitle={International Conference on Learning Representations},
  volume={2025},
  pages={36528--36546},
  year={2025}
}

@inproceedings{zhang2024longcite,
  title={Longcite: Enabling llms to generate fine-grained citations in long-context qa},
  author={Zhang, Jiajie and Bai, Yushi and Lv, Xin and Gu, Wanjun and Liu, Danqing and Zou, Minhao and Cao, Shulin and Hou, Lei and Dong, Yuxiao and Feng, Ling and others},
  booktitle={Findings of the Association for Computational Linguistics: ACL 2025},
  pages={5098--5122},
  year={2025}
}

@inproceedings{shaham2022scrolls,
  title={Scrolls: Standardized comparison over long language sequences},
  author={Shaham, Uri and Segal, Elad and Ivgi, Maor and Efrat, Avia and Yoran, Ori and Haviv, Adi and Gupta, Ankit and Xiong, Wenhan and Geva, Mor and Berant, Jonathan and others},
  booktitle={Proceedings of the 2022 Conference on Empirical Methods in Natural Language Processing},
  pages={12007--12021},
  year={2022}
}

@inproceedings{shaham2023zeroscrolls,
  title={ZeroSCROLLS: A zero-shot benchmark for long text understanding},
  author={Shaham, Uri and Ivgi, Maor and Efrat, Avia and Berant, Jonathan and Levy, Omer},
  booktitle={Findings of the Association for Computational Linguistics: EMNLP 2023},
  pages={7977--7989},
  year={2023}
}

@article{li2023loogle,
  title={Loogle: Can long-context language models understand long contexts?},
  author={Li, Jiaqi and Wang, Mengmeng and Zheng, Zilong and Zhang, Muhan},
  journal={arXiv preprint arXiv:2311.04939},
  year={2023}
}

@inproceedings{dong2023bamboo,
  title={Bamboo: A comprehensive benchmark for evaluating long text modeling capacities of large language models},
  author={Dong, Zican and Tang, Tianyi and Li, Junyi and Zhao, Wayne Xin and Wen, Ji-Rong},
  booktitle={Proceedings of the 2024 Joint International Conference on Computational Linguistics, Language Resources and Evaluation (LREC-COLING 2024)},
  pages={2086--2099},
  year={2024}
}

@inproceedings{zhang2024infinitebench,
  title={$\infty$ Bench: Extending long context evaluation beyond 100K tokens},
  author={Zhang, Xinrong and Chen, Yingfa and Hu, Shengding and Xu, Zihang and Chen, Junhao and Hao, Moo and Han, Xu and Thai, Zhen and Wang, Shuo and Liu, Zhiyuan and others},
  booktitle={Proceedings of the 62nd Annual Meeting of the Association for Computational Linguistics (Volume 1: Long Papers)},
  pages={15262--15277},
  year={2024}
}

@article{li2024needlebench,
  title={NeedleBench: Evaluating LLM Retrieval and Reasoning Across Varying Information Densities},
  author={Li, Mo and Zhang, Songyang and Zhang, Taolin and Duan, Haodong and Liu, Yunxin and Chen, Kai},
  journal={arXiv preprint arXiv:2407.11963},
  year={2024}
}

@article{li2024longiclbench,
  title={Long-context llms struggle with long in-context learning},
  author={Li, Tianle and Zhang, Ge and Do, Quy Duc and Yue, Xiang and Chen, Wenhu},
  journal={arXiv preprint arXiv:2404.02060},
  year={2024}
}

@inproceedings{wang2024loong,
  title={Leave no document behind: Benchmarking long-context llms with extended multi-doc qa},
  author={Wang, Minzheng and Chen, Longze and Cheng, Fu and Liao, Shengyi and Zhang, Xinghua and Wu, Bingli and Yu, Haiyang and Xu, Nan and Zhang, Lei and Luo, Run and others},
  booktitle={Proceedings of the 2024 Conference on Empirical Methods in Natural Language Processing},
  pages={5627--5646},
  year={2024}
}

@inproceedings{bai2024longbenchv2,
  title={Longbench v2: Towards deeper understanding and reasoning on realistic long-context multitasks},
  author={Bai, Yushi and Tu, Shangqing and Zhang, Jiajie and Peng, Hao and Wang, Xiaozhi and Lv, Xin and Cao, Shulin and Xu, Jiazheng and Hou, Lei and Dong, Yuxiao and others},
  booktitle={Proceedings of the 63rd Annual Meeting of the Association for Computational Linguistics (Volume 1: Long Papers)},
  pages={3639--3664},
  year={2025}
}

@article{yen2024helmet,
  title={Helmet: How to evaluate long-context language models effectively and thoroughly},
  author={Yen, Howard and Gao, Tianyu and Hou, Minmin and Ding, Ke and Fleischer, Daniel and Izsak, Peter and Wasserblat, Moshe and Chen, Danqi},
  journal={arXiv preprint arXiv:2410.02694},
  year={2024}
}

@article{miyato2018vat,
  title={Virtual adversarial training: a regularization method for supervised and semi-supervised learning},
  author={Miyato, Takeru and Maeda, Shin-ichi and Koyama, Masanori and Ishii, Shin},
  journal={IEEE transactions on pattern analysis and machine intelligence},
  volume={41},
  number={8},
  pages={1979--1993},
  year={2018},
  publisher={IEEE}
}

@article{tarvainen2017mean,
  title={Mean teachers are better role models: Weight-averaged consistency targets improve semi-supervised deep learning results},
  author={Tarvainen, Antti and Valpola, Harri},
  journal={Advances in neural information processing systems},
  volume={30},
  year={2017}
}

\end{document}